\documentclass{article}

    \PassOptionsToPackage{numbers, compress}{natbib}

 \usepackage[dblblindworkshop, final]{neurips_2025}

\makeatletter
\@workshoptrue
\@ifundefined{@trackname}{}{%
  \renewcommand{\@trackname}{AI for Tabular Data workshop at EurIPS 2025}%
}
\makeatother

\usepackage[utf8]{inputenc} 
\usepackage[T1]{fontenc}    
\usepackage{hyperref}       
\usepackage{url}            
\usepackage{booktabs}       
\usepackage{amsfonts}       
\usepackage{nicefrac}       
\usepackage{subcaption}
\usepackage{microtype}      
\usepackage{xcolor}         
\usepackage{amsmath}
\usepackage{booktabs}
\usepackage{longtable}
\usepackage{graphicx}    
\usepackage{wrapfig}
\usepackage{cleveref}
\usepackage{url}
\usepackage{authblk}



\title{Generalization Can Emerge in Tabular Foundation Models From a Single Table}

\author[1]{Junwei Ma\textsuperscript{$\dagger$}\hspace{0.2em}}
\author[2,3]{Nour Shaheen}
\author[4]{Alex Labach}
\author[2,3]{Amine Mhedhbi}
\author[5,6,7]{Frank Hutter}
\author[4]{\mbox{Anthony L.\ Caterini}}
\author[4]{Valentin Thomas\textsuperscript{$\dagger$}\hspace{0.2em}}

\affil[1]{University of Toronto}
\affil[2]{Polytechnique Montréal}
\affil[3]{Mila – Quebec AI Institute}
\affil[4]{Layer 6 AI}
\affil[5]{Prior Labs}
\affil[6]{ELLIS Institute Tübingen}
\affil[7]{University of Freiburg}

\begin{document}

\maketitle


\begingroup
\renewcommand\thefootnote{\fnsymbol{footnote}}
\footnotetext[2]{Correspondence: \texttt{junweima2@gmail.com}, \texttt{vltn.thomas@gmail.com}}
\endgroup

\vspace{-2em}

\begin{abstract}
\looseness=-1  Deep tabular modelling increasingly relies on in-context learning where, during inference, a model receives a set of $(x,y)$ pairs as context and predicts labels for new inputs without weight updates. 
We challenge the prevailing view that broad generalization here requires pre-training on large synthetic corpora (e.g., TabPFN priors) or a large collection of real data (e.g., TabDPT training datasets), discovering that a relatively small amount of data suffices for generalization.
We find that simple self-supervised pre-training on just a \emph{single} real table can produce surprisingly strong transfer across heterogeneous benchmarks. 
By systematically pre-training and evaluating on many diverse datasets, we analyze what aspects of the data are most important for building a Tabular Foundation Model (TFM) generalizing across domains.
We then connect this to the pre-training procedure shared by most TFMs and show that the number and quality of \emph{tasks} one can construct from a dataset is key to downstream performance.

\end{abstract}

\vspace{-1em}

\section{Introduction}
Tabular data is commonly viewed as highly diverse and heterogeneous, leading to a prevailing belief that transfer learning across different tabular domains is nearly impossible: e.g., \emph{how could training on handwritten digits possibly transfer to estimating California housing prices?} 
However, recent tabular in‑context learning (ICL) methods such as \textsc{TabPFN}~\citep{hollmann2023tabpfn,Hollmann2025}, \textsc{TabDPT}~\citep{ma2025tabdptscalingtabularfoundation}, and \textsc{TabICL}~\citep{qu2025tabicl} -- often termed tabular foundation models (TFMs) -- have challenged this assumption.

\looseness=-1  These methods train transformers using either massive amounts of synthetic data (millions of tables generated from structured priors~\citep{hollmann2023tabpfn,Hollmann2025,qu2025tabicl}) or by sampling from large collections of real tabular datasets~\citep{ma2025tabdptscalingtabularfoundation}, randomizing context composition and prediction targets at each step to encourage in-context generalization.
All state-of-the-art tabular ICL models of which we are aware~\citep{hollmann2023tabpfn,Hollmann2025,ma2025tabdptscalingtabularfoundation, qu2025tabicl, gardner2024large, spinaci2025contexttab} share a similar task construction procedure during pre-training: a dataset is either selected or created, a column of this dataset is used as the target, and a subset of the remaining columns is used as features.
Inference is then performed on new datasets while keeping the pre-trained weights frozen (much like LLMs); labeled examples from the evaluation dataset are given as context and the models must predict the values of unlabeled examples.
The general belief is that, much like in large language models, only a very diverse pre-training set can cover the distribution of unseen tabular tasks and enable out-of-domain generalization.


A surprising finding underpins our work: even with a \emph{single} real-world table -- such as vectorized \textsc{MNIST}~\citep{lecun1998gradient}, where each row represents an image flattened into pixels and each column is a pixel value -- a transformer trained with a na\"ive self-supervised learning (SSL) objective
can still acquire generalization capabilities that transfer robustly \emph{across domains}. 

\begin{figure}[t]
    \centering
    \includegraphics[width=\textwidth]{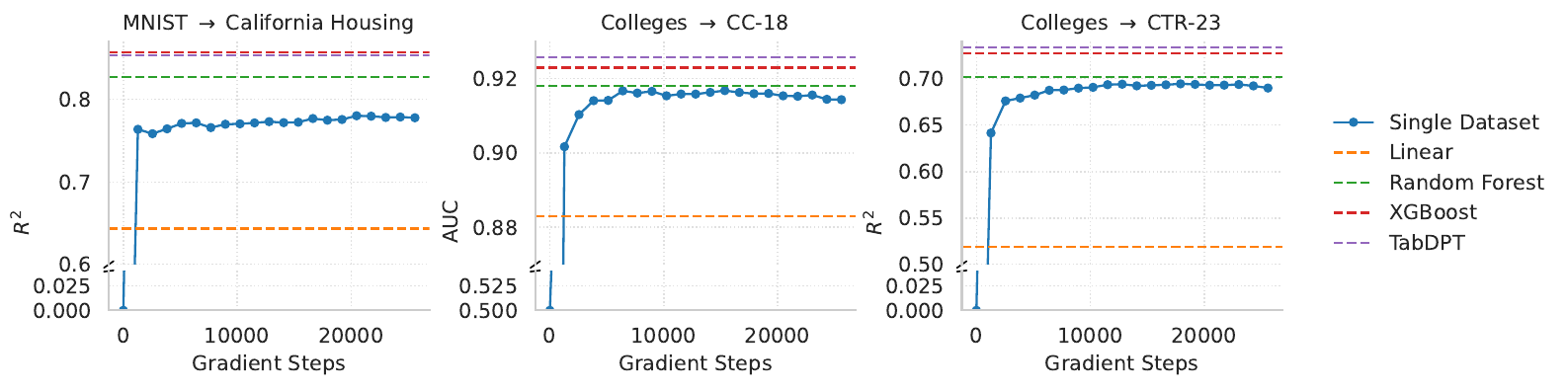}
    \caption{
        Transfer from a single pre‑training dataset. 
        \textbf{Left:} Training only on vectorized \textsc{MNIST} (treated as a table) and evaluating on \textsc{California Housing}.
        \textbf{Middle and Right:} Training on the \textsc{Colleges} dataset and evaluating on the full CC-18 and CTR-23 evaluation suites, respectively.
    }
    \label{fig:transfer_mnist2_california}
\end{figure}

Figure~\ref{fig:transfer_mnist2_california} illustrates this: a transformer trained from scratch on only the \textsc{MNIST} table immediately yields strong in-context performance on structurally and semantically unrelated datasets like \textsc{California Housing}~\citep{pace1997sparse} (predicting real estate prices from geographical and socioeconomic features).
We further train on the \textsc{Colleges} dataset (7k instances, 45 features) \citep{OpenML_dataset_42727}, which contains demographics and institutional features, and evaluate the resulting model on the OpenML CC-18~\citep{bischl2021cc18} (72 datasets) and CTR-23~\citep{fischer2023ctr} (35 datasets) benchmarks.
We find that the model learns quickly and ends up closely matching the performance of a random forest baseline.
This observation stands in contrast to the prevailing assumption that only diverse, massive, or domain-matched pre-training corpora -- such as those used by \textsc{TabPFN} and \textsc{TabDPT} -- could enable such generalization. 



\looseness=-1
In this paper, we investigate which properties of single, real-world datasets for pre-training are most linked to generalization for tabular ICL models. 
We find that: (1) Datasets that transfer well tend to do so universally across evaluation sets; (2) The number of features present in a dataset is a strong indicator of its generalization ability, while the number of instances matters little; and finally (3) The number of \emph{unique tasks} a model is pre-trained on strongly affects its generalization ability.

We believe this work uncovers a very surprising phenomenon and opens the door to more investigations on how much can be learned from even limited amounts of data.


\section{Experiments}

\subsection{Experimental Setup}
We design experiments to explore the central question of what constitutes a ``good dataset'' such that a tabular ICL model pre-trained on it can generalize.
To isolate the effects of the training datasets themselves, we fix both the model architecture and the pre-training procedure throughout most experiments.
Specifically, we use the shared backbone architecture from \textsc{TabPFN}~v1 and \textsc{TabDPT}, a widely adopted design for tabular ICL. 
As for the training objective, we use the pre-training procedure from \textsc{TabDPT} to train tabular ICL models from scratch using a single real dataset only.

To isolate the effect of data, we choose \emph{not} to use retrieval mechanisms (as used in \textsc{TabDPT}) during pre-training, although it is enabled during inference.
We evaluate the generalization performance on the established benchmarks in order to easily compare with the baseline performance directly.

\subsection{Are Good Pre-training Datasets Universally Good?}

\looseness=-1  In~\Cref{fig:transfer_mnist2_california}, we showed the surprising finding that pre-training an ICL tabular model on \textsc{MNIST} can lead to strong performance on other unrelated datasets.
We want to investigate the following question: \emph{``Is each pre-training dataset only good for some downstream task or does it generalize universally?''}

\looseness=-1
Thus, we pre-train models on each of $N_\text{train} = 88$ training datasets (a subset of the \textsc{TabDPT} pre-training data) separately, spanning different modalities and sizes. We evaluated each of the resulting 88 models on $N_\text{eval} = 107$ diverse datasets: the 72 CC-18~\citep{bischl2021cc18} classification datasets and the 35 CTR-23~\citep{fischer2023ctr} regression datasets.
We took special care to ensure no datasets appear in both the pre-training and evaluation corpora.
Linear and RandomForest baselines were computed using \texttt{scikit-learn} defaults\footnote{except with $1000$ maximum iterations for logistic regression instead of $100$}. XGB is the tuned baseline from~\citet{mcelfresh2023neural} and \textsc{TabDPT} is computed from the \href{https://github.com/layer6ai-labs/TabDPT-inference}{public repository}.

\begin{figure}[ht]
    \centering
    \begin{subfigure}[b]{0.45\textwidth}
        \centering
        \includegraphics[width=0.99\textwidth]{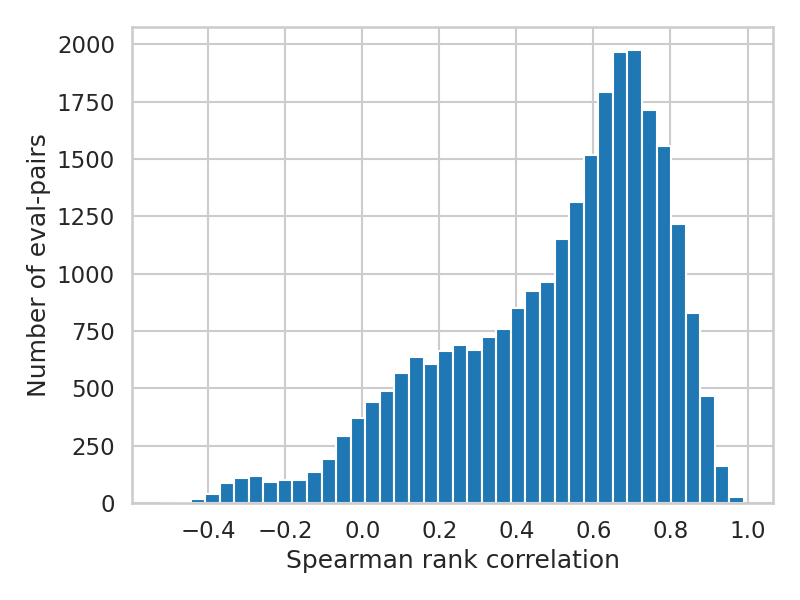}
        \caption{Ranking Correlation: The high values indicate that pre-training datasets which are good on one evaluation dataset tend to be good on others too.}
        \label{fig:stable_ranks}
    \end{subfigure}
    \hfill
    \begin{subfigure}[b]{0.45\textwidth}
        \centering
        \includegraphics[width=0.99\textwidth]{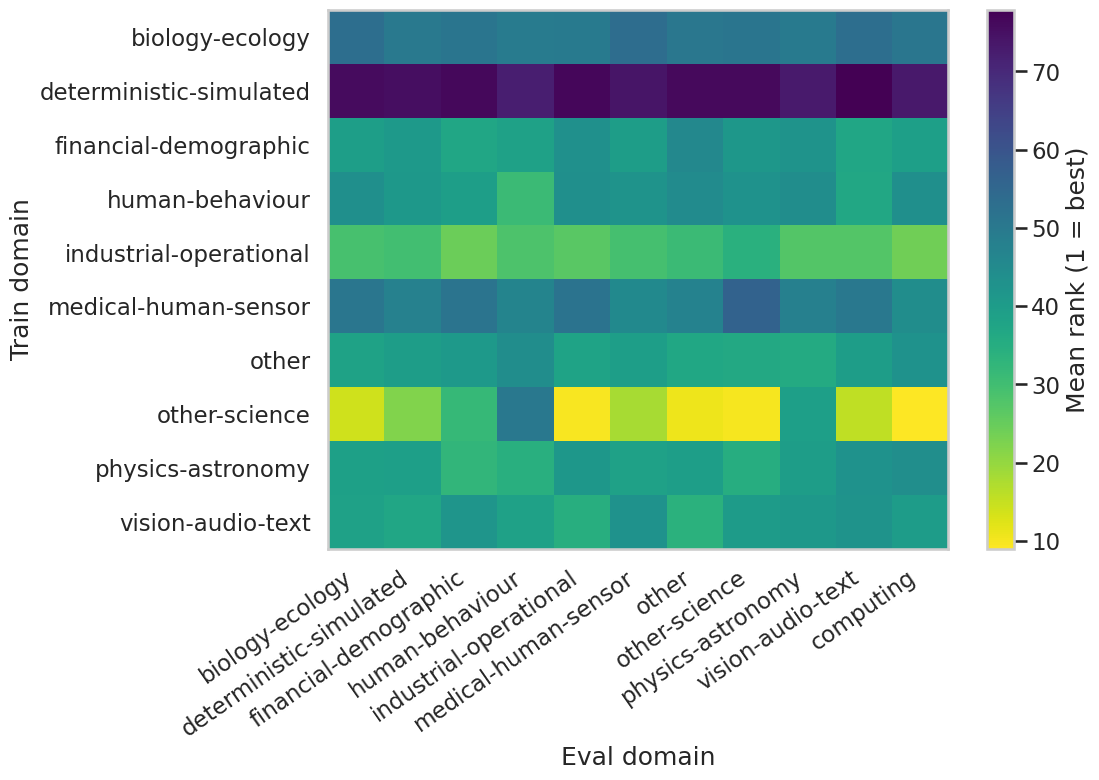}
        \caption{Domain to Domain Transfer: We do not observe stronger transfer when the pre-training and evaluation domain are shared.}
        \label{fig:domain2domain}
    \end{subfigure}
    \caption{Universality of the dataset quality. \textbf{Figure 2a:} We compute the rank of training sets for each evaluation dataset and then plot a histogram of the Spearman correlation between the ranks on all pairs of evaluation datasets. Most evaluation datasets have very correlated ranks. \textbf{Figure 2b:} We group the training and evaluation datasets into distinct domains and plot the density map for average ranks (lower is better). Training and evaluation pairs from the same domain do not appear to transfer better than ones from different domains.}
    \label{fig:universal}
\end{figure}

\looseness=-1
In~\Cref{fig:stable_ranks}, for each evaluation dataset we rank the pre-training datasets based on the performance obtained on the evaluation task, leading to $N_{\text{eval}}$ rankings.
We then gather all pairs of rankings and compute the Spearman correlation. 
Higher correlations indicate that the relative ranking of pre-training datasets is stable.
Since the histogram has much more mass to the right, it
indicates that if a dataset is good for one evaluation dataset it tends to be good for others too, \emph{even across vastly different evaluation tasks}.

\looseness=-1 Next, we investigate whether the \emph{domain} of the single training dataset tends to generalize better to evaluation datasets within the same domain.
For this purpose, we categorize the pre-training and evaluation datasets by manual inspection.
Our protocol is the following: we group the evaluation datasets by domain and take the final performance (either accuracy for classification or correlation for regression) for each single training dataset.
We then rank the training datasets from best to worst for each domain.
Finally we aggregate over the domains of the training datasets.
If datasets from, e.g., Domain A are expected to transfer, on average, better to Domain A, we would expect a strong diagonal dominance in this matrix.
However, in~\Cref{fig:domain2domain} we can see this is absolutely not the case.
Domains do not seem to transfer better to the same domain. Rather, certain domains appear to be better than others across the board (e.g., \texttt{other-science} is great, \texttt{deterministic-simulated} is not).

\subsection{What Constitutes a Good Dataset?}

To understand which properties of a training dataset yield successful generalization, we conduct a meta-analysis using the previous pre-training corpus of $N_\text{train}$ datasets.
For each pre-training dataset, we collect a set of descriptive meta-features, including: the number of features, number of instances, number of categorical features, number of numeric features, amount of missing values, and final pre-training losses for both classification and regression heads.
These meta-features are split into meta-train (80\%) and meta-test (20\%) sets.

\looseness=-1 We then train an XGBoost~\citep{chen2016xgboost} regressor to predict a dataset's average downstream generalization score (chosen as the average of correlation for regression tasks and accuracy for classification tasks) using these meta-features.
The resulting model achieves an $R^2$ of 0.67, indicating that roughly two-thirds of the variance in generalization can be reproduced from straightforward dataset properties alone. 

\begin{figure}[ht]
    \centering
    \begin{subfigure}[b]{0.45\textwidth}
        \centering
    \includegraphics[width=0.99\textwidth]{figures/xgboost\_feature\_importances\_epoch\_100.pdf}
    \caption{
        Feature importance from an XGBoost model trained to predict the average performance across 107 downstream tasks from dataset metadata. 
    }
    \label{fig:xgb_feature_importance}
    \end{subfigure}
    \hfill
    \begin{subfigure}[b]{0.45\textwidth}
        \centering
        \includegraphics[width=0.99\textwidth]{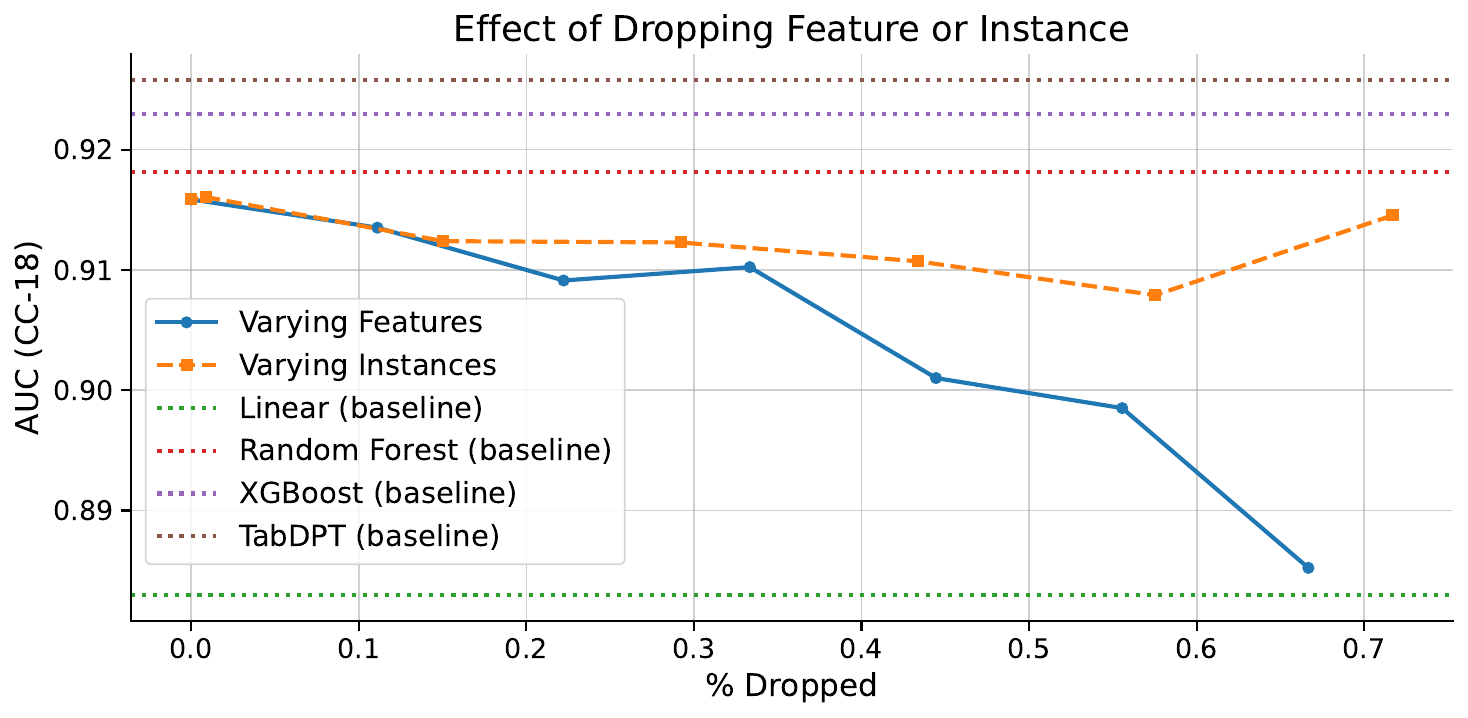}
        \caption{Column vs.\ Row Importance: We analyze the downstream performance when removing either features or instances from the \texttt{College} dataset. 
        }
        \label{fig:feature_instance}
    \end{subfigure}
    \caption{Not all cells in a table are made equal: the number of features matters much more than the number of instances as pre-training dataset for tabular ICL models.}
    \label{fig:universal}
\end{figure}

\looseness=-1
Notably, when evaluating feature importance (see~\Cref{fig:xgb_feature_importance}), the number of features emerges as by far the strongest predictor of downstream transfer performance.
In contrast, the number of instances -- often presumed critical in classical settings -- shows negligible contribution to the XGBoost's predictive power.
This suggests that the richness of the feature space is far more important than purely the dataset size for tabular ICL-based generalization. We validate this on~\Cref{fig:feature_instance} by removing a fraction of either rows or columns on the \textsc{Colleges} dataset, showing that features tend to matter much more than instances. After dropping about 70\% of features, the model performance drops to a similar level as a linear model, while dropping more than 70\% of instances still allows the model to maintain performance close to a random forest. 
We see a similar phenomenon across many other datasets. 


\subsection{Training on Many Good Tasks Unlocks Generalization}

\begin{wrapfigure}{r}{0.52\textwidth} 
  \vspace{-\baselineskip}             
  \centering
  \includegraphics[width=\linewidth]{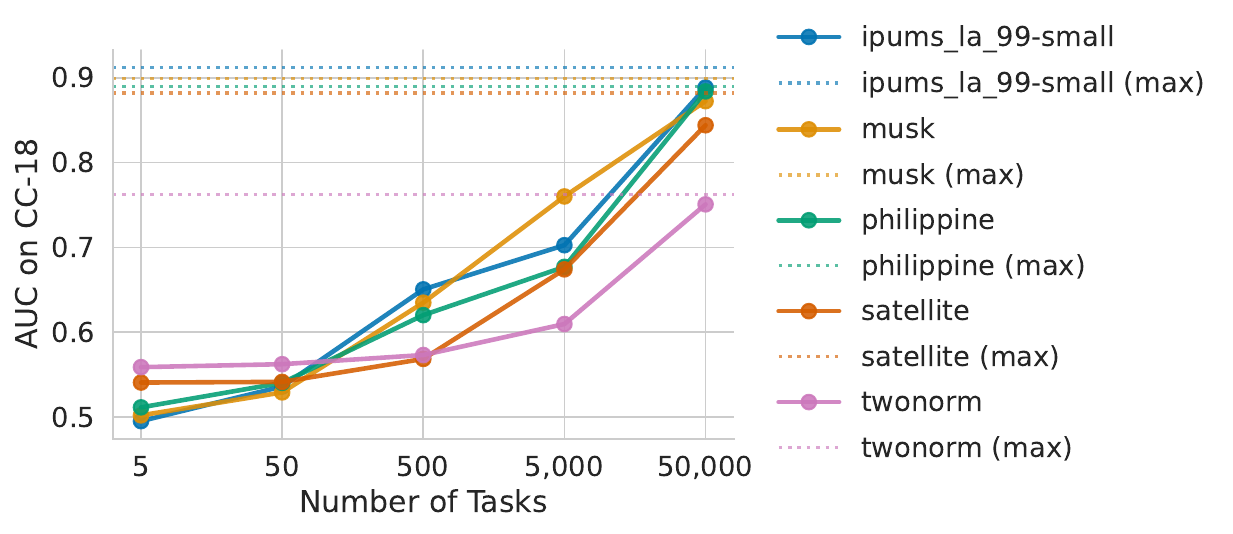}
  \caption{\looseness=-1 Downstream AUC as a function of the number of
  unique tasks used during training. More tasks during pre-training
  consistently leads to better transfer. This demonstrates that both the amount of tasks and the quality of tasks drive the generalization performance.}
  \label{fig:num_tasks_performance}
  \vspace{-\baselineskip}             
\end{wrapfigure}

\looseness=-1 In~\Cref{fig:feature_instance}, we observed that dropping features caused a strong drop in performance. To understand why, we tie this observation back to the randomized procedure used for pre-training. As mentioned, a column is selected as target while a subset of other columns are used as features. We call this a \emph{task}. For instance, a dataset with a fixed supervised target is one task, while using a randomized target and features can have $\mathcal{O}(k\ 2^k)$ tasks for $k$ columns. Thus the role of columns may be more important than the role of rows as it is directly tied to the number of tasks or relationships between features the model is exposed to during training.

We disentangle the impact of the number of tasks from the number of features in~\Cref{fig:num_tasks_performance}. We do not change the amount of instances or features in a given dataset but instead we vary the number of tasks the model is allowed to pre-train on.
To be precise, we select $N$ tasks using allowed permutations (one target, the remaining columns being either masked or used as features) and vary $N$ for several datasets.


Our results in~\Cref{fig:num_tasks_performance} show a striking pattern: across all datasets, increasing the number of tasks from 5 to tens of thousands drives the performance from barely above random ($\text{AUC} = 0.5$) to good performance.
This underscores that while there are important inherent differences across datasets and that some datasets are significantly better than others for pre-training, the main factor explaining generalization across domains is the number of tasks a model was trained on during pre-training.

In other words, it is not so much the amount of data (in tokens or bytes of memory) of the pre-training set that is key, but the amount of tasks / relationships between features and targets trained on that enable a tabular ICL model to generalize. While we believe task diversity and quality are other key elements, we leave a deeper investigation into these factors for future work.

\clearpage

\bibliography{refs}

@article{pace1997sparse,
  title={Sparse Spatial Autoregressions},
  author={Pace, R. Kelley and Barry, Ronald},
  journal={Statistics \& Probability Letters},
  volume={33},
  number={3},
  pages={291--297},
  year={1997},
  publisher={Elsevier}
}

@article{lecun1998gradient,
  title={Gradient-based learning applied to document recognition},
  author={LeCun, Yann and Bottou, L{\'e}on and Bengio, Yoshua and Haffner, Patrick},
  journal={Proceedings of the IEEE},
  volume={86},
  number={11},
  pages={2278--2324},
  year={1998}
}

@inproceedings{hollmann2023tabpfn,
  title={{TabPFN}: A Transformer That Solves Small Tabular Classification Problems in a Second},
  author={Hollmann, Noah and M{\"u}ller, Samuel and Eggensperger, Katharina and Hutter, Frank},
  booktitle={International Conference on Learning Representations},
  year={2023}
}

@inproceedings{bischl2021cc18,
 author = {Bischl, Bernd and Bischl, Bernd and Casalicchio, Giuseppe and Feurer, Matthias and Gijsbers, Pieter and Hutter, Frank and Lang, Michel and Gomes Mantovani, Rafael and van Rijn, Jan and Vanschoren, Joaquin},
 booktitle = {Proceedings of the Neural Information Processing Systems Track on Datasets and Benchmarks},
 title = {{OpenML} Benchmarking Suites},
 volume = {1},
 year = {2021}
}

@inproceedings{fischer2023ctr,
  title={{OpenML-CTR23} -- {A} curated tabular regression benchmarking suite},
  author={Fischer, Sebastian Felix and Feurer, Matthias and Bischl, Bernd},
  booktitle={AutoML Conference (Workshop)},
  year={2023}
}

@inproceedings{chen2016xgboost,
  title={{XGBoost}: A scalable tree boosting system},
  author={Chen, Tianqi and Guestrin, Carlos},
  booktitle={Proceedings of the 22nd ACM SigKDD International Conference on Knowledge Discovery and Data Mining},
  year={2016}
}

@misc{OpenML_dataset_42727,
  author = {Gijsbers, Pieter},
  title = {Colleges},
  year = {2020},
  howpublished = {OpenML Dataset 42727},
  url = {https://www.openml.org/d/42727},
  note = {Accessed: 2025-10-21}
}

@inproceedings{mcelfresh2023neural,
  title={When do neural nets outperform boosted trees on tabular data?},
  author={McElfresh, Duncan and Khandagale, Sujay and Valverde, Jonathan and Prasad C, Vishak and Ramakrishnan, Ganesh and Goldblum, Micah and White, Colin},
  booktitle={Advances in Neural Information Processing Systems},
  year={2023}
}

@inproceedings{spinaci2025contexttab,
  title     = {{ConTextTab}: A Semantics-Aware Tabular In-Context Learner},
  author    = {Marco Spinaci and Marek Polewczyk and Maximilian Schambach and Sam Thelin},
  booktitle = {1st ICML Workshop on Foundation Models for Structured Data},
  year      = {2025}
}

@inproceedings{qu2025tabicl,
  title={{TabICL}: A Tabular Foundation Model for In-Context Learning on Large Data},
  author={Qu, Jingang and Holzm{\"u}ller, David and Varoquaux, Ga{\"e}l and Morvan, Marine Le},
  booktitle={International Conference on Machine Learning},
  year={2025}
}

@inproceedings{gardner2024large,
  title={Large Scale Transfer Learning for Tabular Data via Language Modeling},
  author={Gardner, Josh and Perdomo, Juan C and Schmidt, Ludwig},
  booktitle={Advances in Neural Information Processing Systems},
  year={2024}
}

@article{Hollmann2025,
  author = {Hollmann, Noah and Müller, Samuel and Purucker, Lennart and Krishnakumar, Arjun and Körfer, Max and Hoo, Shi Bin and Schirrmeister, Robin Tibor and Hutter, Frank},
  title = {Accurate Predictions on Small Data with a Tabular Foundation Model},
  journal = {Nature},
  volume = {637},
  number = {8045},
  pages = {319--326},
  year = {2025}
}

@inproceedings{ma2025tabdptscalingtabularfoundation,
      title={{TabDPT}: Scaling Tabular Foundation Models on Real Data}, 
      author={Junwei Ma and Valentin Thomas and Rasa Hosseinzadeh and Hamidreza Kamkari and Alex Labach and Jesse C. Cresswell and Keyvan Golestan and Guangwei Yu and Anthony L. Caterini and Maksims Volkovs},
      year={2025},
        booktitle={Advances in Neural Information Processing Systems}, 
}

\clearpage


\appendix

\section{Appendix / supplemental material}

\subsection{Impact of Pre-training Dataset on CC-18 and CTR-23 Performance}

Tables\ref{tab:cc18_auc}, \ref{tab:ctr23_r2} report the performance of TabDPT when pre-trained on different datasets and evaluated on two benchmark suites: \textbf{CC-18} for classification (measured by AUC) and \textbf{CTR-23} for regression (measured by R\textsuperscript{2}). Each table lists the results for all pre-training datasets, along with their number of instances, their number of features, and their domain. The results are sorted in descending order of performance, highlighting which pre-training datasets achieve better performance on each benchmark.

\begin{longtable}{l r r r l}
\caption{AUC scores on CC-18 for TabDPT pre-trained on various datasets, sorted in descending order of AUC.}
\label{tab:cc18_auc}\\
\toprule
\textbf{Dataset Name} & \textbf{AUC on CC-18} & \textbf{\# Instances} & \textbf{\# Features} & \textbf{Dataset Domain} \\
\midrule
\endhead

\bottomrule
\endfoot
colleges &  0.916 &  7063 &  45 &  other \\
ipums\_la\_97-small &  0.914 &  7019 &  61 &  financial-demographic \\
cardiotocography &  0.913 &  2126 &  36 &  medical-human-sensor \\
ipums\_la\_98-small &  0.912 &  7485 &  56 &  financial-demographic \\
ipums\_la\_99-small &  0.912 &  8844 &  57 &  financial-demographic \\
volkert &  0.910 &  58310 &  181 &  other \\
KDDCup09\_appetency &  0.910 &  50000 &  231 &  human-behaviour \\
APSFailure &  0.909 &  76000 &  171 &  industrial-operational \\
KDDCup09\_churn &  0.909 &  50000 &  231 &  industrial-operational \\
Census-Income &  0.907 &  299285 &  42 &  financial-demographic \\
gas-drift &  0.906 &  13910 &  129 &  other-science \\
gas-drift-different-concentrations &  0.903 &  13910 &  130 &  other-science \\
kick &  0.903 &  72983 &  33 &  industrial-operational \\
MiniBooNE &  0.902 &  130064 &  51 &  physics-astronomy \\
christine &  0.901 &  5418 &  1637 &  other \\
road-safety &  0.901 &  111762 &  33 &  human-behaviour \\
dilbert &  0.900 &  10000 &  2001 &  other \\
musk &  0.899 &  6598 &  168 &  other-science \\
covertype &  0.898 &  581012 &  55 &  biology-ecology \\
PizzaCutter3 &  0.894 &  1043 &  38 &  other \\
scene &  0.894 &  2407 &  300 &  vision-audio-text \\
dionis &  0.893 &  416188 &  61 &  other \\
eye\_movements &  0.893 &  10936 &  28 &  medical-human-sensor \\
helena &  0.892 &  65196 &  28 &  other \\
philippine &  0.890 &  5832 &  309 &  other \\
heloc &  0.889 &  10000 &  23 &  financial-demographic \\
SpeedDating &  0.888 &  8378 &  121 &  human-behaviour \\
one-hundred-plants-texture &  0.887 &  1599 &  65 &  biology-ecology \\
kdd\_internet\_usage &  0.886 &  10108 &  69 &  financial-demographic \\
fbis.wc &  0.886 &  2463 &  2001 &  vision-audio-text \\
jannis &  0.883 &  83733 &  55 &  other \\
colleges\_usnews &  0.883 &  1302 &  34 &  other \\
Satellite &  0.882 &  5100 &  37 &  physics-astronomy \\
porto-seguro &  0.881 &  595212 &  58 &  human-behaviour \\
PieChart3 &  0.881 &  1077 &  38 &  other \\
sylva\_agnostic &  0.879 &  14395 &  217 &  biology-ecology \\
jasmine &  0.874 &  2984 &  145 &  other \\
sylvine &  0.874 &  5124 &  21 &  other \\
elevators &  0.871 &  16599 &  19 &  other \\
guillermo &  0.870 &  20000 &  4297 &  other \\
JapaneseVowels &  0.868 &  9961 &  15 &  vision-audio-text \\
ada &  0.867 &  4147 &  49 &  other \\
cjs &  0.865 &  2796 &  35 &  biology-ecology \\
higgs &  0.865 &  98050 &  29 &  physics-astronomy \\
ada\_agnostic &  0.862 &  4562 &  49 &  financial-demographic \\
shuttle &  0.854 &  58000 &  10 &  physics-astronomy \\
house\_16H &  0.850 &  22784 &  17 &  financial-demographic \\
pol &  0.849 &  10082 &  27 &  industrial-operational \\
okcupid-stem &  0.840 &  50789 &  20 &  human-behaviour \\
eeg-eye-state &  0.839 &  14980 &  15 &  medical-human-sensor \\
analcatdata\_halloffame &  0.833 &  1340 &  17 &  other \\
page-blocks &  0.828 &  5473 &  11 &  vision-audio-text \\
default-of-credit-card-clients &  0.825 &  13272 &  21 &  financial-demographic \\
spoken-arabic-digit &  0.825 &  263256 &  15 &  vision-audio-text \\
albert &  0.823 &  425240 &  79 &  other \\
MagicTelescope &  0.820 &  19020 &  12 &  physics-astronomy \\
mushroom &  0.820 &  8124 &  23 &  biology-ecology \\
magic &  0.819 &  19020 &  11 &  physics-astronomy \\
Click\_prediction\_small &  0.816 &  39948 &  12 &  human-behaviour \\
pbcseq &  0.815 &  1945 &  19 &  medical-human-sensor \\
fabert &  0.814 &  8237 &  801 &  other \\
sf-police-incidents &  0.807 &  2215023 &  9 &  human-behaviour \\
credit &  0.807 &  16714 &  11 &  financial-demographic \\
ldpa &  0.802 &  164860 &  8 &  medical-human-sensor \\
airlines &  0.794 &  539383 &  8 &  industrial-operational \\
artificial-characters &  0.786 &  10218 &  8 &  deterministic-simulated \\
ada\_prior &  0.768 &  4562 &  15 &  financial-demographic \\
madeline &  0.765 &  3140 &  260 &  other \\
twonorm &  0.762 &  7400 &  21 &  deterministic-simulated \\
house\_8L &  0.758 &  22784 &  9 &  financial-demographic \\
Diabetes130US &  0.756 &  71090 &  8 &  medical-human-sensor \\
yeast &  0.748 &  1269 &  9 &  biology-ecology \\
pollen &  0.748 &  3848 &  6 &  biology-ecology \\
hill-valley &  0.744 &  1212 &  101 &  deterministic-simulated \\
walking-activity &  0.737 &  149332 &  5 &  medical-human-sensor \\
compas-two-years &  0.737 &  4966 &  12 &  human-behaviour \\
visualizing\_soil &  0.727 &  8641 &  5 &  biology-ecology \\
poker-hand &  0.706 &  1025009 &  11 &  deterministic-simulated \\
puma8NH &  0.698 &  8192 &  9 &  deterministic-simulated \\
kropt &  0.688 &  28056 &  7 &  deterministic-simulated \\
chess &  0.686 &  28056 &  7 &  deterministic-simulated \\
kr-vs-k &  0.675 &  28056 &  7 &  deterministic-simulated \\
analcatdata\_supreme &  0.668 &  4052 &  8 &  other \\
\end{longtable}

\begin{longtable}{l r r r l}
\caption{R\textsuperscript{2} scores on CTR-23 for TabDPT pre-trained on various datasets, sorted in descending order of R\textsuperscript{2}.}\label{tab:ctr23_r2}\\
\toprule
\textbf{Dataset Name} & \textbf{R\textsuperscript{2} on CTR-23}& \textbf{\# Instances} & \textbf{\# Features}  & \textbf{Dataset Domain}  \\
\midrule
\endhead

\bottomrule
\endfoot
colleges &  0.694 &  7063 &  45 &  other \\
covertype & 0.679 & 581012 & 55 & biology-ecology \\
ipums\_la\_99\-small & 0.677 & 8844 & 57 & financial-demographic \\
ipums\_la\_98\-small & 0.672 & 7485 & 56 & financial-demographic \\
kick & 0.668 & 72983 & 33 & industrial-operational \\
Census-Income & 0.666 & 299285 & 42 & financial-demographic \\
gas-drift-different-concentrations & 0.666 & 13910 & 130 & other-science \\
volkert & 0.663 & 58310 & 181 & other \\
ipums\_la\_97\-small & 0.662 & 7019 & 61 & financial-demographic \\
KDDCup09\_appetency & 0.661 & 50000 & 231 & human-behaviour \\
MiniBooNE & 0.661 & 130064 & 51 & physics-astronomy \\
APSFailure & 0.660 & 76000 & 171 & industrial-operational \\
KDDCup09\_churn & 0.659 & 50000 & 231 & industrial-operational \\
gas-drift & 0.656 & 13910 & 129 & other-science \\
cardiotocography & 0.656 & 2126 & 36 & medical-human-sensor \\
scene & 0.652 & 2407 & 300 & vision-audio-text \\
jannis & 0.652 & 83733 & 55 & other \\
dilbert & 0.652 & 10000 & 2001 & other \\
philippine & 0.652 & 5832 & 309 & other \\
helena & 0.650 & 65196 & 28 & other \\
one-hundred-plants-texture & 0.644 & 1599 & 65 & biology-ecology \\
PizzaCutter3 & 0.644 & 1043 & 38 & other \\
eye\_movements & 0.642 & 10936 & 28 & medical-human-sensor \\
sylva\_agnostic & 0.642 & 14395 & 217 & biology-ecology \\
dionis & 0.641 & 416188 & 61 & other \\
colleges\_usnews & 0.640 & 1302 & 34 & other \\
christine & 0.638 & 5418 & 1637 & other \\
road-safety & 0.636 & 111762 & 33 & human-behaviour \\
heloc & 0.633 & 10000 & 23 & financial-demographic \\
PieChart3 & 0.633 & 1077 & 38 & other \\
JapaneseVowels & 0.626 & 9961 & 15 & vision-audio-text \\
higgs & 0.625 & 98050 & 29 & physics-astronomy \\
SpeedDating & 0.624 & 8378 & 121 & human-behaviour \\
sylvine & 0.620 & 5124 & 21 & other \\
elevators & 0.619 & 16599 & 19 & other \\
shuttle & 0.613 & 58000 & 10 & physics-astronomy \\
musk & 0.612 & 6598 & 168 & other-science \\
jasmine & 0.608 & 2984 & 145 & other \\
analcatdata\_halloffame & 0.606 & 1340 & 17 & other \\
Satellite & 0.605 & 5100 & 37 & physics-astronomy \\
page-blocks & 0.604 & 5473 & 11 & vision-audio-text \\
pol & 0.602 & 10082 & 27 & industrial-operational \\
eeg-eye-state & 0.595 & 14980 & 15 & medical-human-sensor \\
house\_16H & 0.592 & 22784 & 17 & financial-demographic \\
guillermo & 0.581 & 20000 & 4297 & other \\
porto-seguro & 0.576 & 595212 & 58 & human-behaviour \\
spoken-arabic-digit & 0.562 & 263256 & 15 & vision-audio-text \\
cjs & 0.558 & 2796 & 35 & biology-ecology \\
fbis.wc & 0.553 & 2463 & 2001 & vision-audio-text \\
pbcseq & 0.551 & 1945 & 19 & medical-human-sensor \\
ada & 0.546 & 4147 & 49 & other \\
albert & 0.543 & 425240 & 79 & other \\
ada\_agnostic & 0.542 & 4562 & 49 & financial-demographic \\
artificial-characters & 0.541 & 10218 & 8 & deterministic-simulated \\
MagicTelescope & 0.540 & 19020 & 12 & physics-astronomy \\
kdd\_internet\_usage & 0.536 & 10108 & 69 & financial-demographic \\
mushroom & 0.533 & 8124 & 23 & biology-ecology \\
magic & 0.528 & 19020 & 11 & physics-astronomy \\
ldpa & 0.510 & 164860 & 8 & medical-human-sensor \\
okcupid-stem & 0.510 & 50789 & 20 & human-behaviour \\
default-of-credit-card-clients & 0.496 & 13272 & 21 & financial-demographic \\
ada\_prior & 0.490 & 4562 & 15 & financial-demographic \\
credit & 0.476 & 16714 & 11 & financial-demographic \\
madeline & 0.465 & 3140 & 260 & other \\
visualizing\_soil & 0.459 & 8641 & 5 & biology-ecology \\
compas-two-years & 0.452 & 4966 & 12 & human-behaviour \\
fabert & 0.446 & 8237 & 801 & other \\
Click\_prediction\_small & 0.430 & 39948 & 12 & human-behaviour \\
airlines & 0.421 & 539383 & 8 & industrial-operational \\
sf-police-incidents & 0.410 & 2215023 & 9 & human-behaviour \\
yeast & 0.405 & 1269 & 9 & biology-ecology \\
house\_8L & 0.384 & 22784 & 9 & financial-demographic \\
hill-valley & 0.383 & 1212 & 101 & deterministic-simulated \\
pollen & 0.329 & 3848 & 6 & biology-ecology \\
twonorm & 0.322 & 7400 & 21 & deterministic-simulated \\
Diabetes130US & 0.313 & 71090 & 8 & medical-human-sensor \\
walking-activity & 0.312 & 149332 & 5 & medical-human-sensor \\
analcatdata\_supreme & 0.296 & 4052 & 8 & other \\
kropt & 0.290 & 28056 & 7 & deterministic-simulated \\
kr-vs-k & 0.276 & 28056 & 7 & deterministic-simulated \\
chess & 0.275 & 28056 & 7 & deterministic-simulated \\
puma8NH & 0.271 & 8192 & 9 & deterministic-simulated \\
poker-hand & 0.263 & 1025009 & 11 & deterministic-simulated \\
\end{longtable}

\end{document}